\documentclass{article}
\usepackage{spconf,amsmath,graphicx}

\usepackage{subfigure}
\usepackage{graphicx}

\usepackage{indentfirst}
\usepackage{latexsym}

\usepackage{dsfont}
\usepackage{txfonts}
\usepackage{amsfonts}

\title{Fractal Dimension Pattern Based Multiresolution Analysis for Rough Estimator of Person-Dependent Audio Emotion Recognition}
\twoauthors
  {Miao Cheng}
	{Department of Computer Science\\
	Qingdao University, Qingdao, China\\
    E-mail: mew.cheng@gmail.com}
  {Ah Chung Tsoi}
	{Faculty of Information Technology\\
	Macau University of Science and Technology, Macau\\
	E-mail: actsoi@must.edu.mo}

\begin{document}
%
\maketitle
\begin{abstract}

As a general means of expression, audio analysis and recognition has attracted much attentions for its wide applications in real-life world. Audio emotion recognition (AER) attempts to understand emotional states of human with the given utterance signals, and has been studied abroad for its further development on friendly human-machine interfaces.
In this work, the discriminant patterns of audio emotions are conducted as multiresolution analysis of utterance signals, and fractal dimension features are calculated for acoustic feature extraction. Furthermore, it is able to efficiently learn intrinsic characteristics of auditory emotions, while the utterance features are learned from fractal dimensions of each sub-bands.
Experimental results show the proposed method is able to provide comparative performance for audio emotion recognition.


\end{abstract}

\section{Introduction}

Audio emotion analysis aims to understand the intrinsic explanation of emotional states of human. As a common sense, audio emotion recognition (AER) identifies the emotional information from speech signals, and learns audio features for classification of different emotions. Furthermore, audio analysis and recognition has become more and more important and attractive with advancement on wide applications of mobile devices, and brought much perspective of applicable reality of life demand. And it has been found useful for many applications, such as human-machine conversation \cite{Watanabe90hmc}, emotion understanding \cite{Maaoui08ER}, sickness diagnosis \cite{Syed06diagnosis}, and so on.
In the literature, AER has been studied abroad on different aspects. One of these outstanding solutions, is to exploit the differential features from audio data, and find the discriminative information for identification of different emotions.

Without loss of generality, there have been \emph{six} emotions, such as \emph{angry}, \emph{disgust}, \emph{fear}, \emph{happy}, \emph{sad}, and \emph{surprise}, involved in most emotion recognition works. Until now, several audio features has been widely adopted to depict the audio information from a perspective point of distinctive classification.
With the most significant characteristics of affect in speech, the pitch is usually estimated based on the Fourier analysis of the logarithmic amplitude spectrum of the signal \cite{Hoch05pitch}, which is divided into a set of frames by windowing. As a low level of features, pitch is used to describe the acoustic signals for its utilised statistics. The zero crossing rate \cite{Gouyon00zcr} calculates average result of the number of times that the audio signal crosses zero within a particular time window, which is to determine the audio patterns of fevered and agitated emotions as a pattern feature.

The log-energy features \cite{Zeng04logEng} are able to find the distinctive patterns of certain emotions among different audio signals, and decide the emotional states with the amplitude of a segment of speech. And as we found, it can observably differentiate negative emotions for specific discriminant learning as reachable numeric parameters, while a stable performance is available. Distinguishingly, the teager energy operator (TEO) \cite{Gao07teo} uses the nonlinear operator to measure the changing energies of nonlinear emotional components. The well-known Mel-Frequency Cepstral Coefficients (MFCCs) \cite{Kwon03mfcc} is based on the short-term power spectrum of speech signals, and a linear cosine transform of a log power spectrum on a nonlinear Mel scale of frequency. Then, the coefficients are collected via the obtained mel-frequency cepstrum (MFC) for approximating to the human auditory system's response.
Nevertheless, it is always hardly to determine the best choice of representative features for specific audio data, and insufficiently to correctly classify emotions with single features.


\begin{figure*}
\centering
\subfigure[Pitch]{\includegraphics[width=.475\textwidth]{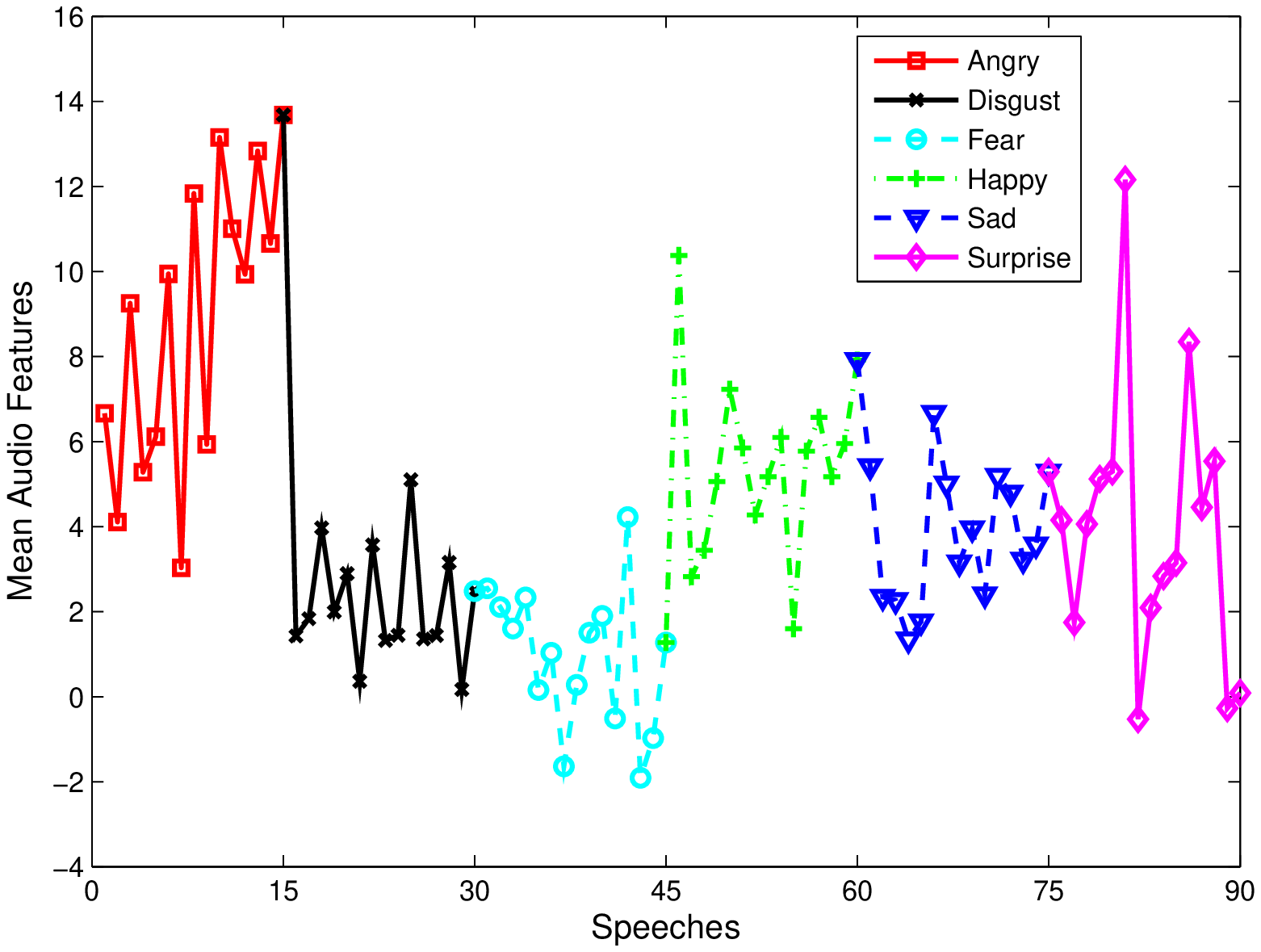}}
\subfigure[Zero crossing rates]{\includegraphics[width=.475\textwidth]{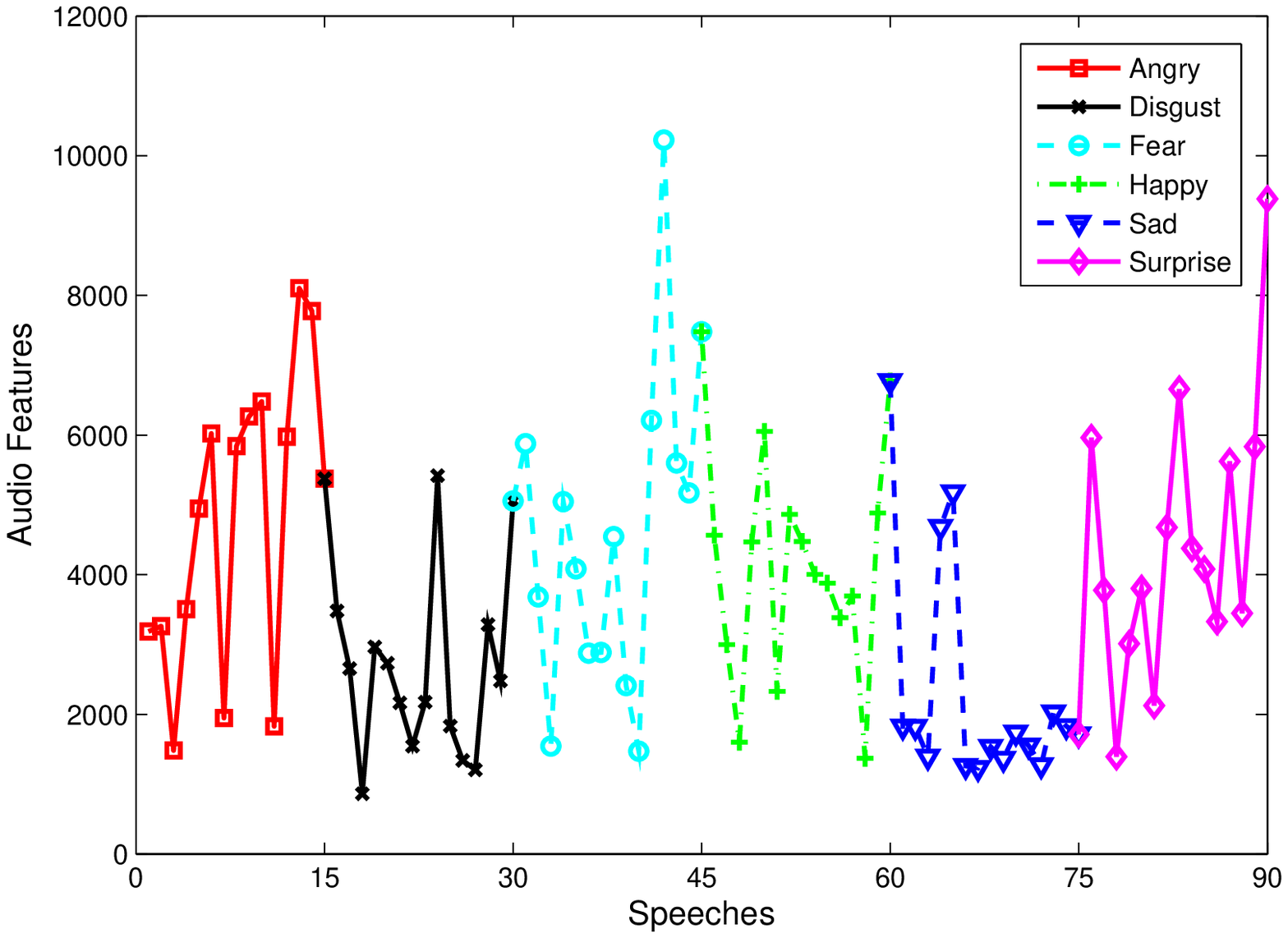}}
\caption{The audio features from one individual of SAVEE data set. (a) Pitch features, (b) Zero crossing rates.}
\end{figure*}
\section{Wavelet Transform and Fractal Dimension}
As a development of traditional Fourier transform, wavelet transform (WT) has been widely applied to signal processing and pattern analysis \cite{Boggess09Wavelet}. With multi-scale decomposition idea, the outstanding advantage of WT is its complete multi-scale analysis of signal inputs,  while approximation is reached in light of approximate and detail coefficients  \cite{Daubechies92Wavelet}. Actually, WT can be explained as an approximate the original signal with the scaling functions and wavelet functions.
Generally, multi-scale analysis constructs a set of nested function space, namely $ \cdots \subset { V }_{ -2 }\subset { V }_{ -1 }\subset { V }_{ 0 }\subset { V }_{ 1 }\subset { V }_{ 2 }\cdots  $. Therefore, the differential subtraction of neighboring function spaces defines a space formed by wavelet functions, namely $ { V }_{ j }\bigoplus { W }_{ j }={ V }_{ j+1 } $. In addition, wavelet packet based tree-structure \cite{Wang15TFF} has also been widely applied to AER for its universally applicable property.


On the other hand, Fractal dimension (FD) is a kind of nonlinear approximate methods of complicated measure, and has been widely applied to pattern recognition and bioinformatics \cite{Yang15FSWTPS}. For certain practical measures, it is difficult to exploit the metrical results directly. As a result, the original problem is usually reduced to a count of fractal units, and outlets are collected to reach the total measure as the so-called fractal dimension. In fact, several FD methods have been designed to conduct different kinds of calculational demands, such as Hausdorff dimension \cite{Felix18FD}, box counting dimension \cite{Falconer14FD}, Katz's dimension \cite{Katz88FD}, and Higuchi's dimension \cite{Higuchi88FD}.

The Hausdorff dimension \cite{Felix18FD} measures the local size of a space taking into account the distance between points, the metric. For shapes that are smooth, or shapes with a small number of corners, the shapes of traditional geometry and science, the Hausdorff dimension is an integer agreeing with the topological dimension. In fractal geometry, box-counting dimension \cite{Falconer14FD} is a way of determining the fractal dimension of a set S in a Euclidean space $ \mathds{R}^n $, which is also known as Minkowski dimension. In addition, there are also some FD holding the ability of conducting the sequence data for deterministic metric of signal sample. More specifically, Katz's FD \cite{Katz88FD} is calculated as:
\begin{equation}
 D=\frac { log\left( L/a \right)  }{ log\left( { d }/{ a } \right)  }  = \frac { log\left( n \right)  }{ log\left( n \right) + log\left( { d }/{ L } \right)  },
\end{equation}
where $ L $ and $ a $ denote the sum and average of the Euclidean distances between the successive point of the sample as well as the maximum distance between the first point and any other point of the sample $ d $. Given an one-dimensional signal $ x_i, \left( i = 1, \cdots, N \right) $, Higuchi's FD \cite{Higuchi88FD} firstly calculates sub-sample sets from the signal data as
\begin{equation}
\begin{array}{ll}
 X_k^m & = \left\{ {X\left( {m + ik} \right)} \right\}_{i = 0}^{\left[ {{{\left( {N - m} \right)} \mathord{\left/
 {\vphantom {{\left( {N - m} \right)} k}} \right.
 \kern-\nulldelimiterspace} k}} \right]}\\
 {} & = \left\{ {x\left( m \right),x\left( {m + k} \right),x\left( {m + 2k} \right)}, \right.\\
 {} & \cdots \begin{array}{*{20}{c}}
{\left. {x\left( {m + \left[ {\frac{{N - m}}{k}} \right]k} \right)} \right\},}&{m = 1, \cdots, k}.
\end{array}
\end{array}
\end{equation}
Here, $ k \in \left[ {1,{k_{\max }}} \right] $ and $ N $ is the sample size. Then the approximated length of each $ X_k $ can be calculated as
\begin{equation}
 {L_m}\left( k \right) = \frac{{\sum\limits_{i = 1}^{\left[ {\frac{{N - m}}{k}} \right]} {\left| {X\left( {m + ik} \right) - X\left( {m + \left( {i - 1} \right)k} \right)} \right|\left( {N - 1} \right)} }}{{\left[ {{{\left( {N - m} \right)} \mathord{\left/
 {\vphantom {{\left( {N - m} \right)} k}} \right.
 \kern-\nulldelimiterspace} k}} \right]{k^2}}}.
\end{equation}
Finally, the fractal dimension of data is obtained by solving the problem
\begin{equation}
 \left\langle {L\left( k \right)} \right\rangle  \propto {k^{ - D}}.
\end{equation}

\begin{figure*}
\centering
\subfigure[Log Energy]{\includegraphics[width=.475\textwidth]{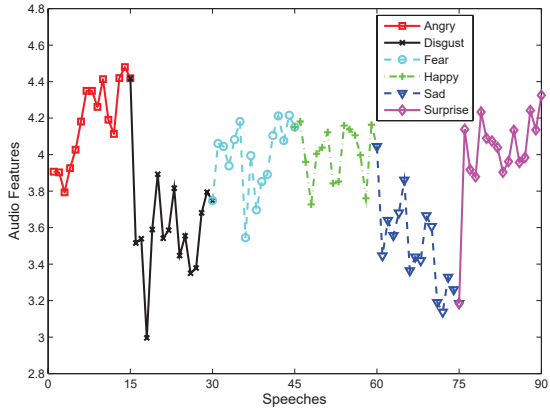}}
\subfigure[TEO]{\includegraphics[width=.475\textwidth]{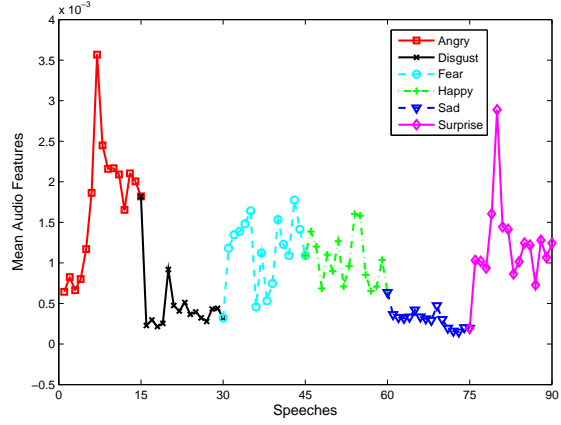}}
\caption{The audio features from one individual of SAVEE data set. (a) Log Engery, (b) TEO.}
\end{figure*}
\section{Fractal Dimension based Audio Emotion Recognition}
For an AER system, it is necessary to exploit the discriminative patterns with respect to chosen audio features. And an acceptable results usually depend on suitable features that best represent the distinctive characteristics of specific data, which has been a common sense for audio analysis. In this work, the SAVEE data set
\footnote[1]{http://kahlan.eps.surrey.ac.uk/savee/}
is employed for audio emotion analysis, and resultant outlets refer to audio feature extraction as disclosed. As a difficult data set for AER, it is hardly to exploit the audio patterns between audio data of different individuals. To make the rules clear, several audio features extracted from SAVEE data set are illustrated in Fig. 1 and Fig. 2.

\begin{figure}
\centering
\subfigure[DC]{\includegraphics[width=.22\textwidth]{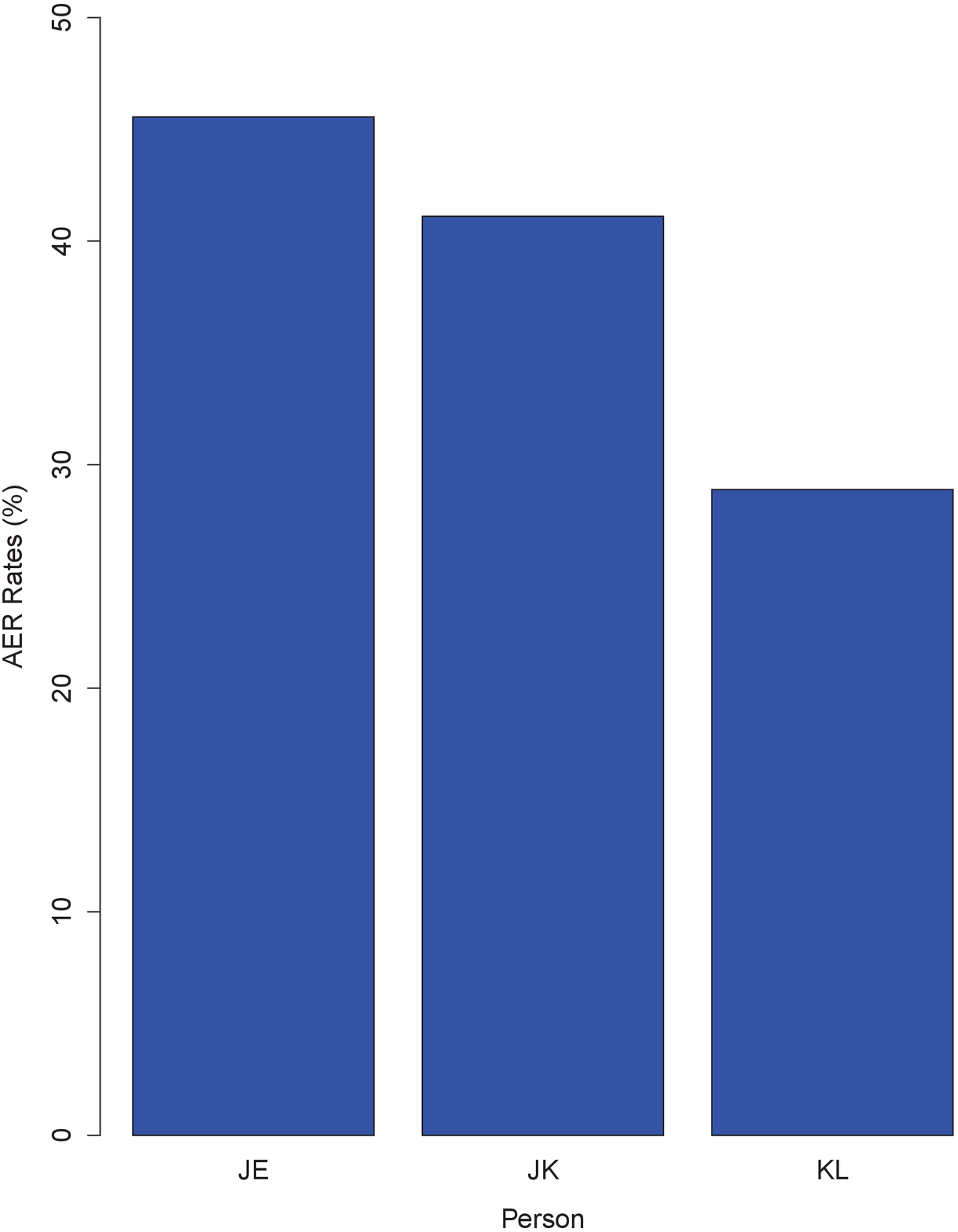}}
\subfigure[JE]{\includegraphics[width=.22\textwidth]{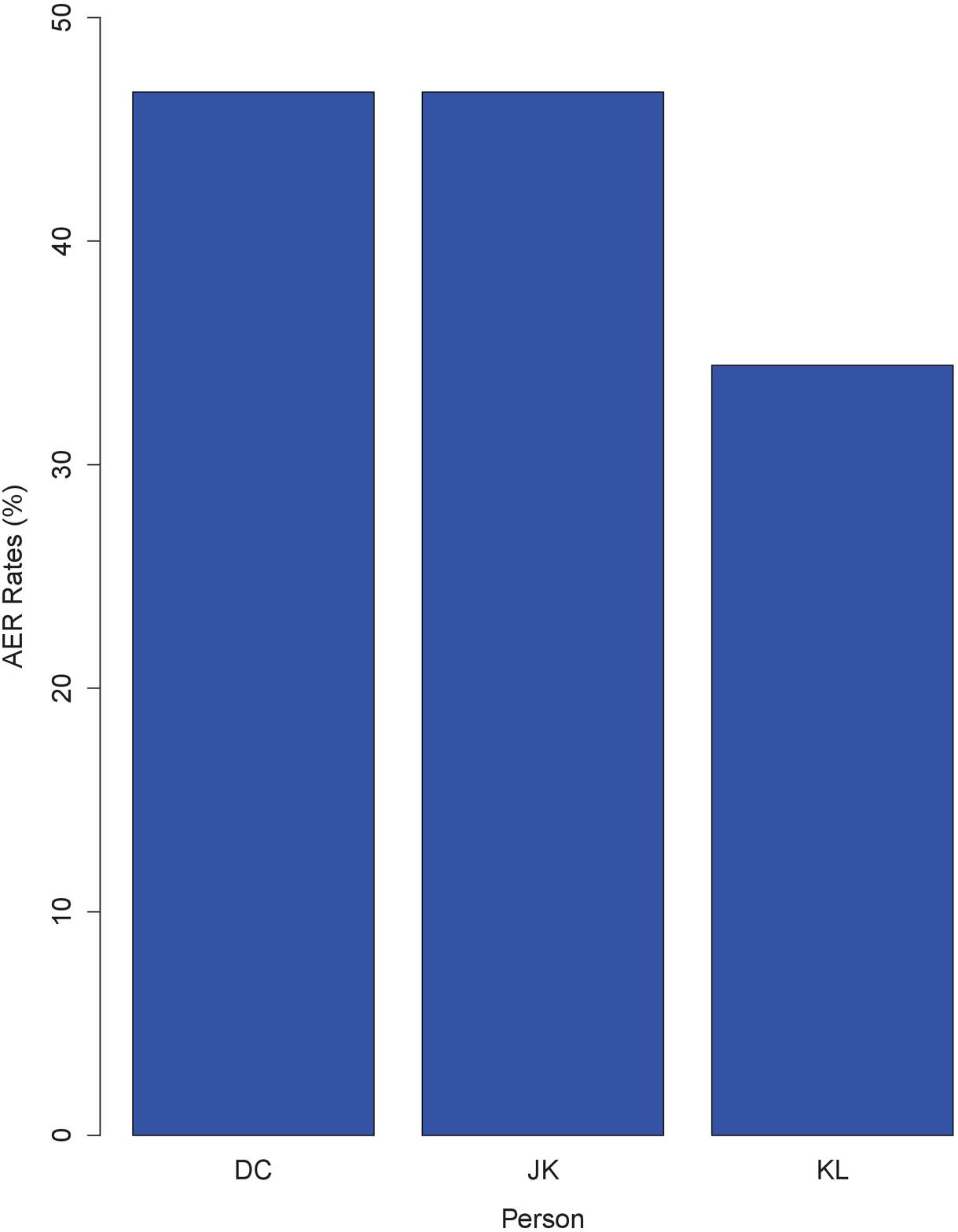}}
\subfigure[JK]{\includegraphics[width=.22\textwidth]{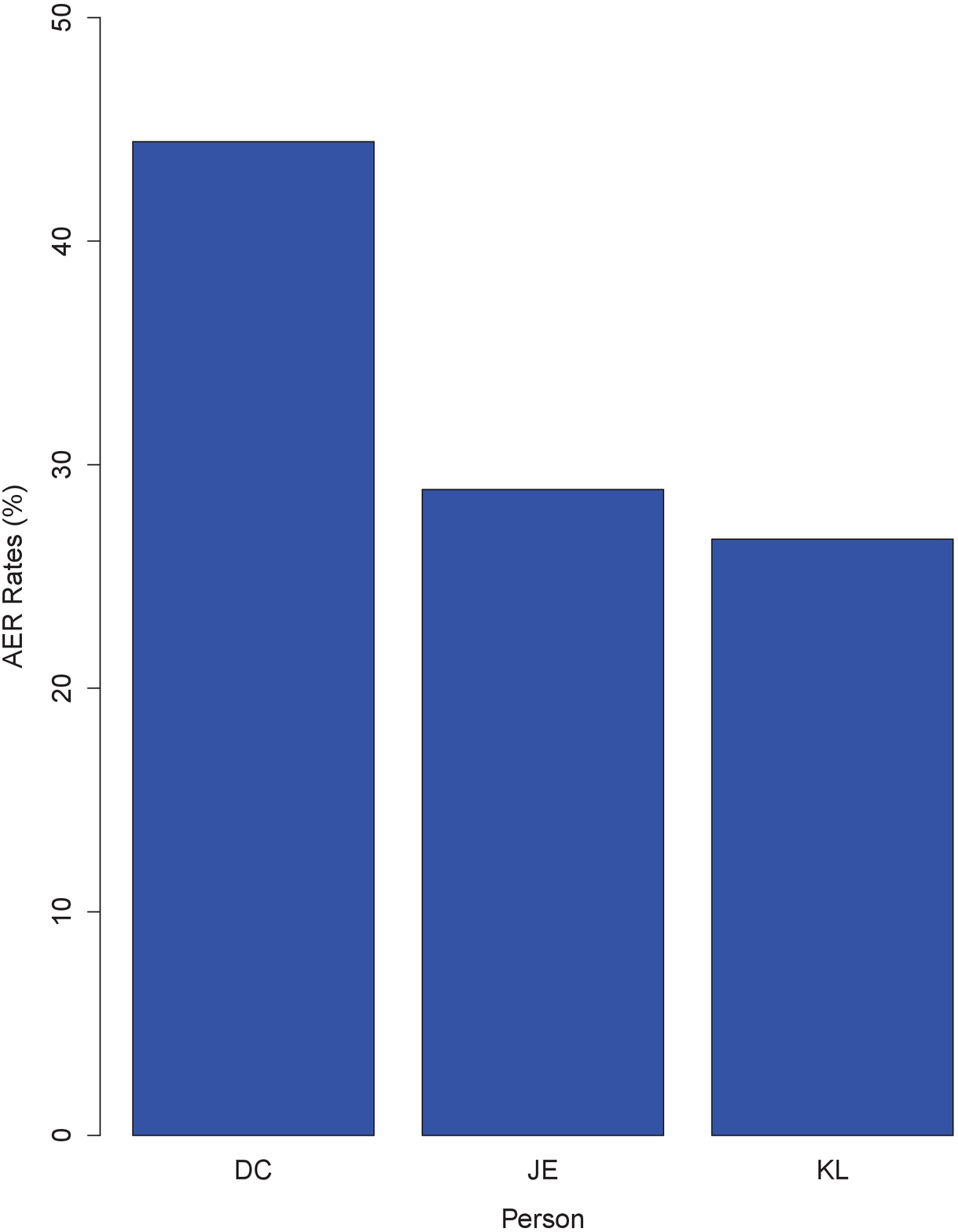}}
\subfigure[KL]{\includegraphics[width=.22\textwidth]{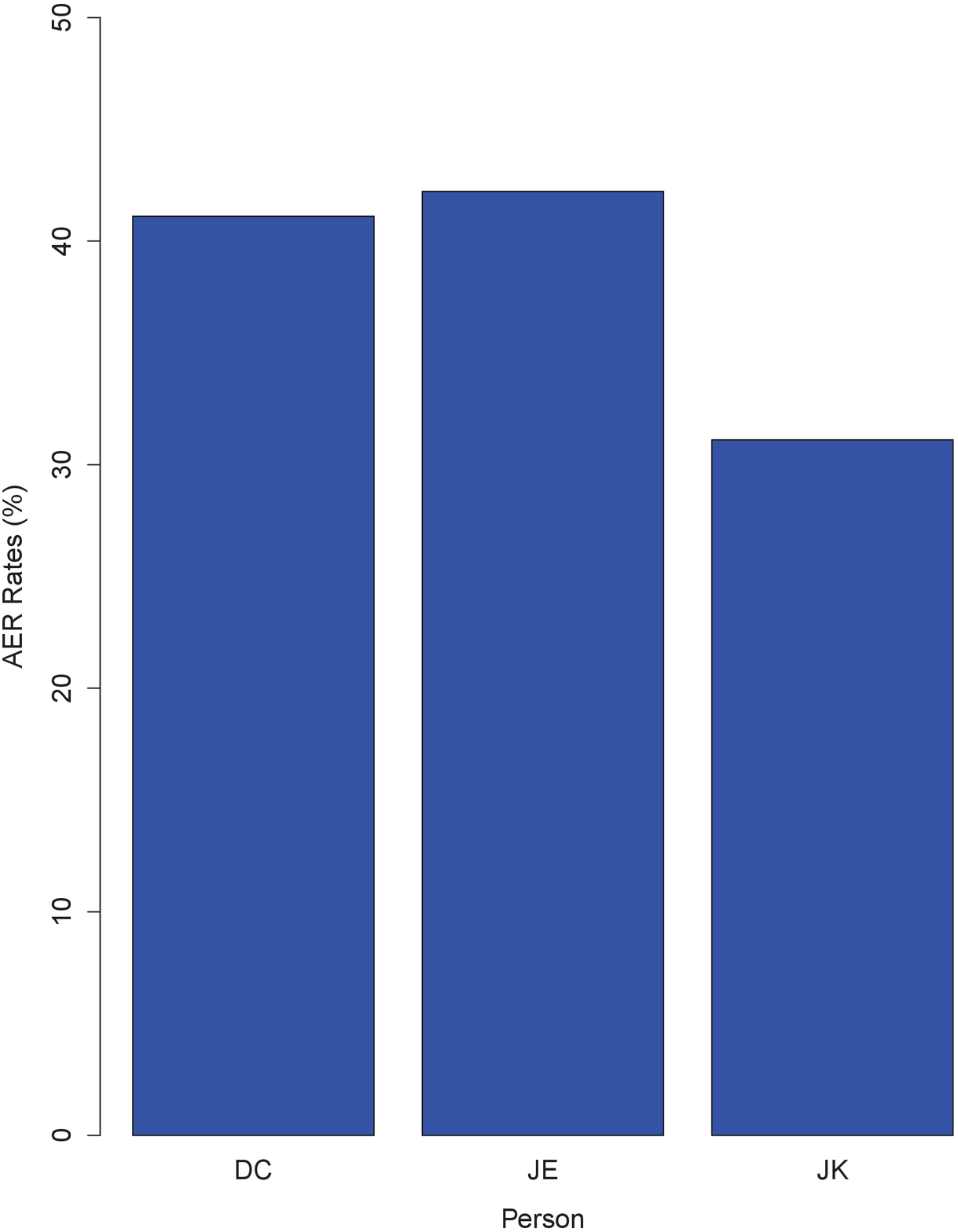}}
\caption{The AER results on SAVEE data set of one vs. three individuals for learning. The results corresponding to the labeled information of one individual: (a) DC, (b) JE, (c) JK, (d) KL.}
\end{figure}
As illustrated, the pitch features and zero crossing rates are hardly to depict the distinctive differences among different emotions, especially for emotional features of \emph{disgust}, \emph{fear}, \emph{happy}, \emph{sad}. Nevertheless, sad and disgust emotions can be selected with much lower energies over others in most cases, while \emph{angry} emotion gives obvious information associated with higher TEO features among all kinds of speeches. In light of this consideration, it is feasible to discriminant different emotions according to these distinctive features step by step. Thereafter, the audio features of each individual can be divided into several portions of instances corresponding to Energy and TEO features firstly. And \emph{disgust}, \emph{sad} and \emph{angry} among all emotions are able to be sequentially screened out in principle. Though it is always necessary to learn the discriminative patterns during each screening steps, however, it  indeed reduces the complexities of emotion classification problems. Different from some existing works that refer to a screening steps for AER only \cite{Ooi14AER}, it still fails to reach good results of complicated audio patterns, e.g., SAVEE. Actually, it may give the worst results for correct classification of audio features, if no mechanism else has involved. To address this limitation, FD features are borrowed in depiction of audio emotions, and concrete response to utterance signals can be obtained.

In terms of definition of FD, the sequential data can be represented as lists of FD features, and discriminant patterns can be handled in following steps. As described foregoing, there are several FD methods that can be adopted to sample data, and it is optional for AER in general. Nevertheless, Higuchi's FD is picked up in this work, due to its simple implementation and its better performance over other methods as testing. Furthermore, the inconsistent lengths of features can be avoided, and FD features are extracted in each wavelet decomposition levels. Though there are also several works conducting feature extraction in a similar manner, it is worthwhile to highlight the differences here:
\begin{itemize}
\item There is no floating windows or fussy iterations, e.g., dictionary learning \cite{Rubinstein98sparse}, involved in discriminative matching, which has been a popular solution widely adopted in many related works. And thus, the efficient achievement of AER is able to be preserved.
\item It takes both approximate and detail coefficients of each level for feature extraction, though many solutions only refer to the approximate coefficients for discriminative learning. And it is found discriminant ability can be improved with detailed pattern information.
\end{itemize}


\begin{figure}
\centering
\subfigure[DC and JE]{\includegraphics[width=.22\textwidth]{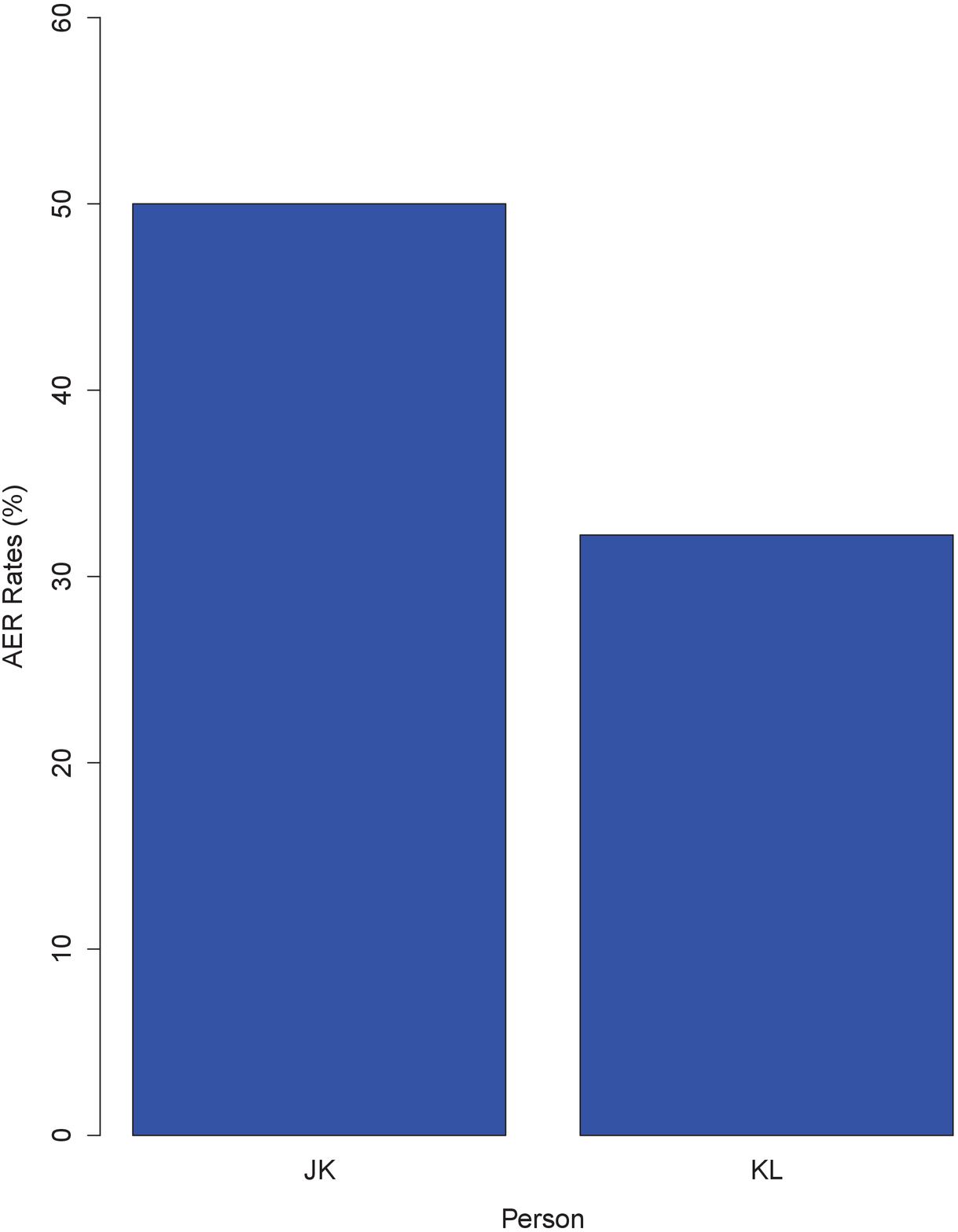}}
\subfigure[DC and KL]{\includegraphics[width=.22\textwidth]{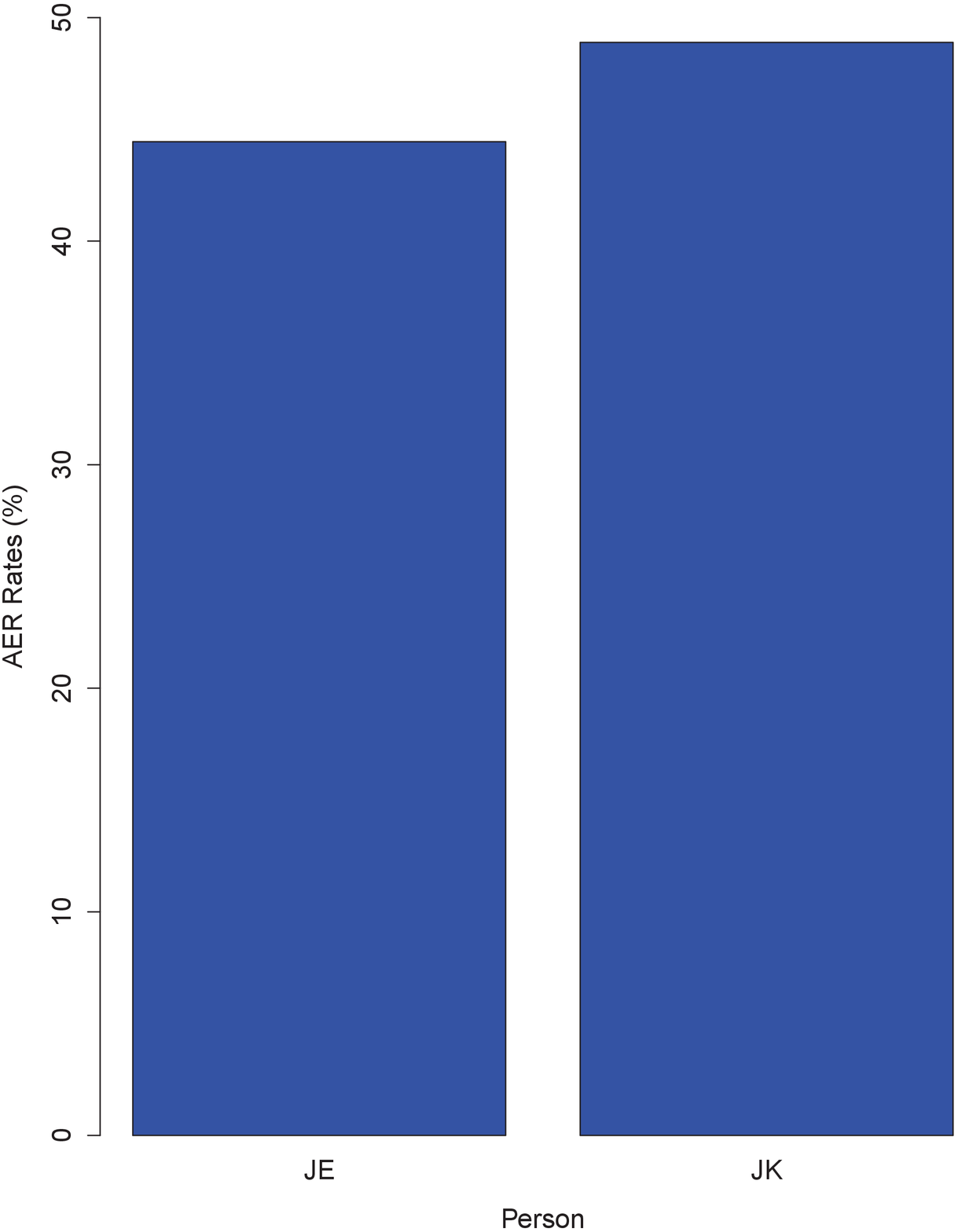}}
\subfigure[JE and JK]{\includegraphics[width=.22\textwidth]{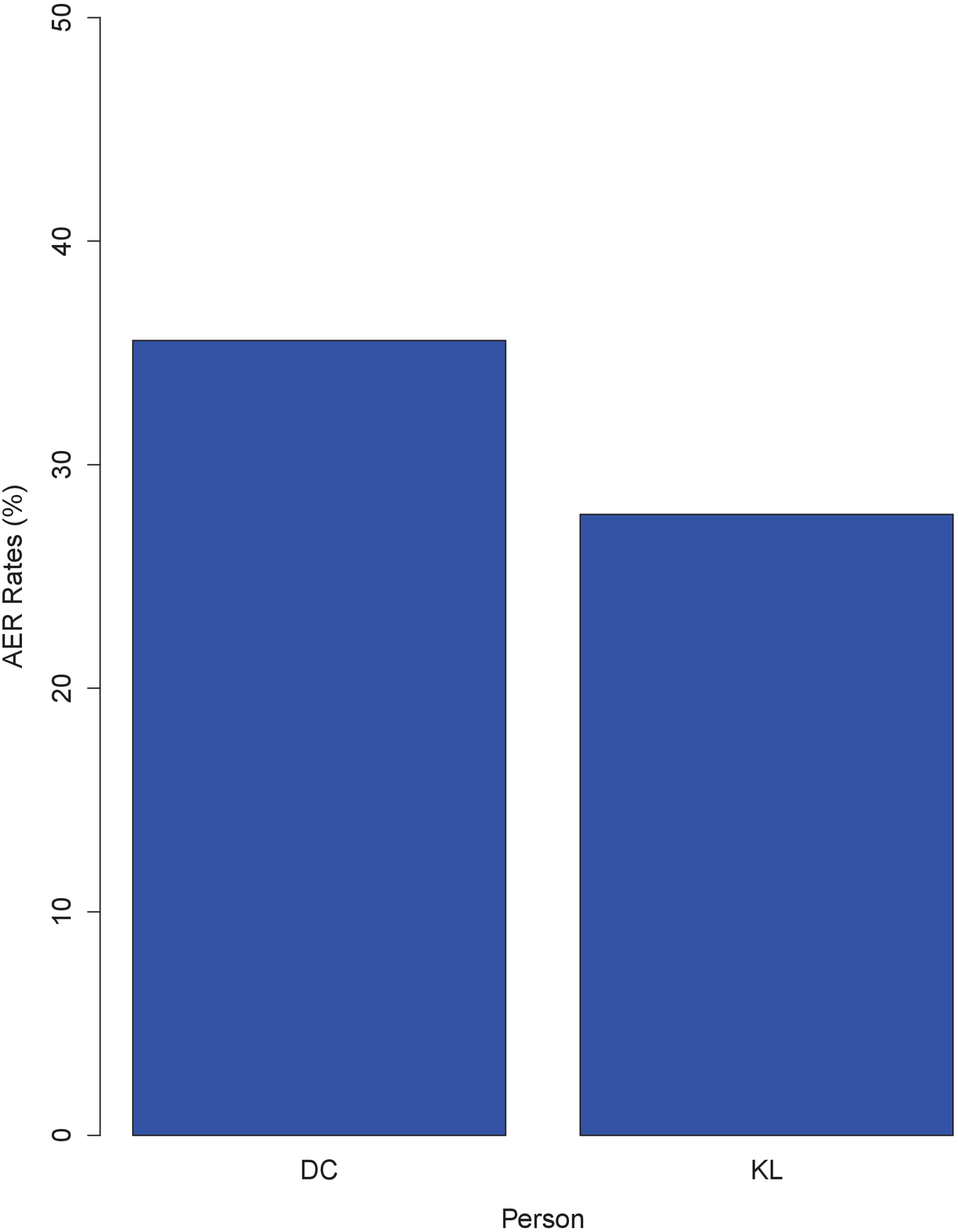}}
\subfigure[JK and KL]{\includegraphics[width=.22\textwidth]{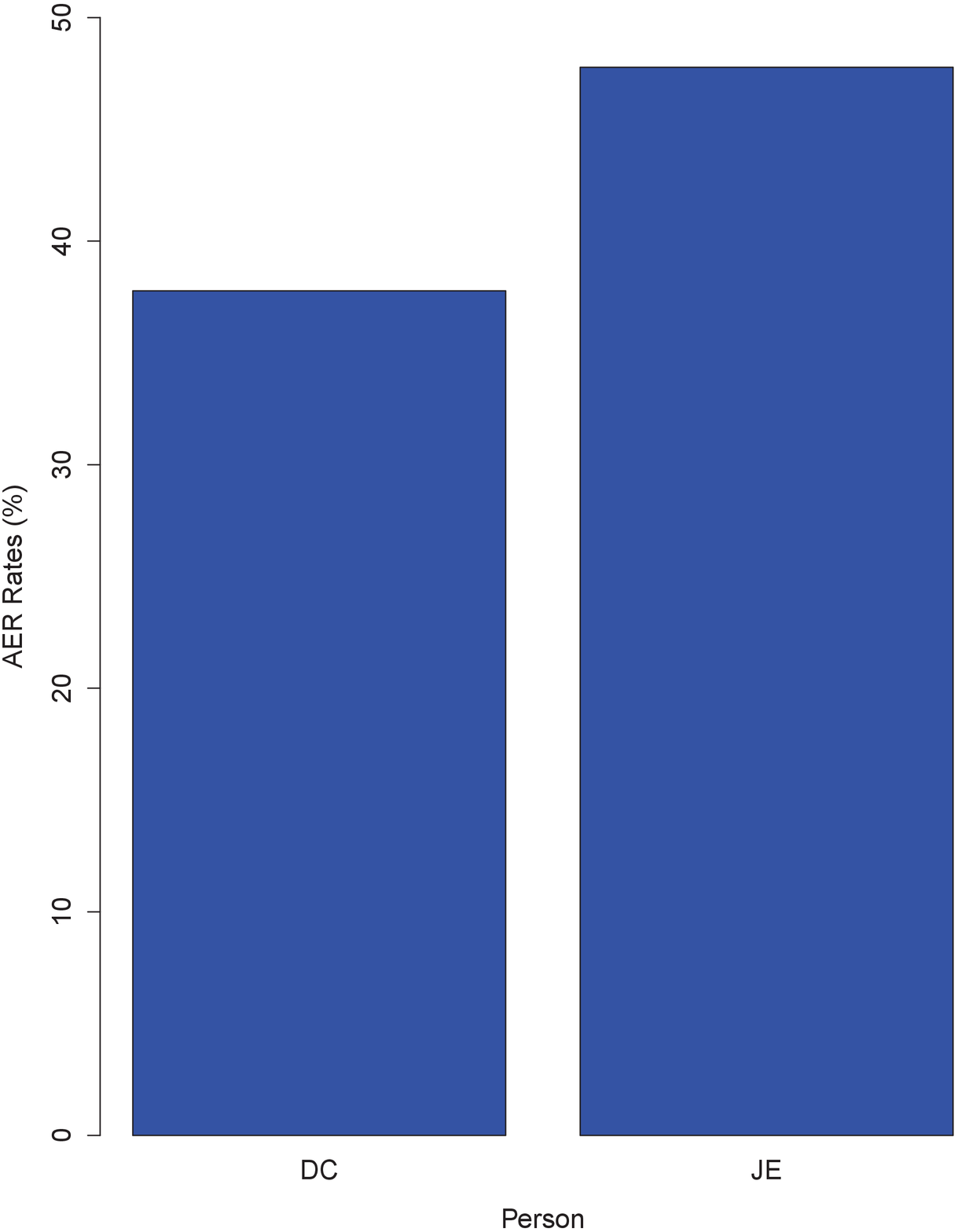}}
\caption{The AER results on SAVEE data set of two vs. two individuals for learning. The results corresponding to the labeled information of two individuals: (a) DC and JE, (b) DC and KL, (c) JE and JK, (d) JK and KL.}
\end{figure}
\section{Experimental Results}
In this section, the experimental performance of proposed method on AER are evaluated, and discriminant performance is disclosed with respect to different individuals. The Surrey Audio-Visual Expressed Emotion (SAVEE) data set 
consists of recordings from 4 male actors in 7 different emotions, 480 British English utterances in total. And it has been well-known for its difficulty of accurate recognition of emotions with respect to each data modality. In our experiments, the audio data of six standard emotions are used, while the \emph{neutral} emotion is ignored, and then there are 90 utterances for each individual. To learn the reduced features, the maximum margin criterion (MMC) \cite{Miao11MMC} associated with KNN classifier is employed for discriminative classification.

Firstly, the audio data of one actor is picked up for learning, while the rest utterances of other individuals are all for testing of AER. That is, the audio data of one vs. three individuals  are set up for learning, and the divisions of utterances are proceeded with respect to audio features of each individual. The AER rates corresponding to each referred person is shown in Fig. 3. In terms of the results, it is difficult to learn the ideal auditory features no matter whose audio data is used for labeled information. Nevertheless, the best results are able to reach an approximate to 45\% accuracy,  on behalf of FD features and screening step, and it is obvious that KL is the hardest one for AER in every cases. Oppositely, the utterances of JK is the easiest one for auditory emotion capture, whose stable recognition results can be obtained for each referred person. In addition, the AER results of JE is also acceptable due to its similar utterance characteristics as JK.

In the second experiments, the audio data from two individuals are selected for pattern analysis, while the rest data of two other persons are used for emotion recognition. The experimental results are illustrated in Fig. 4. As shown, the recognition performance can be improved in most cases of choices of referred audio data, and around 50\% accuracy can be reached. Nevertheless, there are still some obtained results consistent with the foregoing ones generated from few available patterns, due to certain hard data sets for accurate classification. And the outstanding results come from the discriminative learning of final linear classifier, benefit from the fact of more labeled data are available for pattern analysis.

\section{Conclusion}
In this work, AER is considered as a multi-scale analysis problem, and FD features are calculated to represent the best distinctive patterns of different audio emotions. Distinguishingly, both approximate and detail coefficients of wavelet decomposition have been adopted to conduct the hard emotion data. Furthermore, no floating window is used in the signal decomposition stage, while emotion portions are also enjoined for further reduced learning. Experimental results show that the proposed method is able to contribute comparative performance for AER, even if the audio emotion is quite distinctive from each other of individuals for recognition.

\bibliographystyle{IEEEbib}
\bibliography{AERef}

\begin{thebibliography}{10}

\bibitem{Watanabe90hmc}
T.~Watanabe,
\newblock ``The adaptation of machine conversational speed to speaker utterance
  speed in human-machine communication,''
\newblock {\em IEEE Trans. Sys. Man Cybern.}, vol. 20, no. 2, pp. 502--507,
  1990.

\bibitem{Maaoui08ER}
C.~Maaoui, A.~Pruski, and F.~Abdat,
\newblock ``Emotion recognition for human-machine communication,''
\newblock in {\em Proc. Int. Conf. Intell. Robots and Sys.}, 2008.

\bibitem{Syed06diagnosis}
Z.~Syed, D.~Leeds, D.~Curtis, J.~Guttag, F.~Nesta, and R.~A. Levine,
\newblock ``Audio-visual tools for computer-assisted diagnosis of cardiac
  disorders,''
\newblock in {\em Proc. Int. Symp. Comp. Med. Syst.}, 2006, pp. 207--212.

\bibitem{Hoch05pitch}
S.~Hoch, F.~Althoff, G.~McGlaun, and G.~Rigoll,
\newblock ``Bimodal fusion of emotinal data in an automotive environment,''
\newblock in {\em Proc. IEEE Int. Conf. on Acoustic, Speech, and Signal
  Processing}, 2005, pp. 1085--1088.

\bibitem{Gouyon00zcr}
F.~Gouyon, F.~Pachet, and O.~Delete,
\newblock ``Classifying percussive sounds: a matter of zero-crossing rate?,''
\newblock in {\em Proc. COST G-6 Conf. Digital Audio Effects}, 2001, pp.
  53--58.

\bibitem{Zeng04logEng}
Z.~Zeng,
\newblock ``Bimodal hci-related affect recognition,''
\newblock in {\em Proc. Int. Conf. Multimodal Interfaces}, 2004.

\bibitem{Gao07teo}
H.~Gao, S.~Chen, and G.~Su,
\newblock ``Emotion classification of mandarin speech based on teo nonlinear
  features,''
\newblock in {\em ACIS Int. Conf. Soft. Eng., Artif. Intell. Networking, and
  Parallel/Distributed Computing}, 2007.

\bibitem{Kwon03mfcc}
O.~Kwon, K.~Chan, J.~Hao, and T.~Lee,
\newblock ``Emotion recognition by speech signals,''
\newblock in {\em Proc. Euro. Conf. on Speech Comm. and Tech.}, 2003.

\bibitem{Boggess09Wavelet}
A.~Boggess and F.~J. Narcowich,
\newblock {\em A First Course in Wavelets with Fourier Analysis},
\newblock Wiley, 2nd edition, 2009.

\bibitem{Daubechies92Wavelet}
Ingrid Daubechies,
\newblock {\em Ten Lectures on Wavelets},
\newblock SIAM, 1992.

\bibitem{Wang15TFF}
K.~Wang,
\newblock ``Time-frequency feature representation using multi-resolution
  texture analysis and acoustic activity detector for real-life speech emotion
  recognition,''
\newblock {\em Sensors}, vol. 15, no. 1, pp. 1458--1478, 2015.

\bibitem{Yang15FSWTPS}
L.~Yang, Y.~Y. Tang, Y.~Lu, and H.~Luo,
\newblock ``A fractal dimension and wavelet transform based method for protein
  sequence similarity analysis,''
\newblock {\em IEEE/ACM Trans. Compu. Bio. and Bioinfo.}, vol. 12, no. 2, pp.
  348--359, 2015.

\bibitem{Felix18FD}
H.~Felix,
\newblock ``Dimension und \"{a}u{\ss}eres ma{\ss},''
\newblock {\em Mathematische Annalen}, vol. 79, no. 1-2, pp. 157--179, 1918.

\bibitem{Falconer14FD}
Kenneth Falconer,
\newblock {\em Fractal Geometry: Mathematical Foundations and Applications},
\newblock Wiley, third edition edition, 2014.

\bibitem{Katz88FD}
M.~J. Katz,
\newblock ``Fractals and the analysis of waveforms,''
\newblock {\em Computers in Biology and Medicine}, vol. 18, no. 3, pp.
  145--156, 1988.

\bibitem{Higuchi88FD}
T.~Higuchi,
\newblock ``Approach to an irregular time series on the basis of the fractal
  theory,''
\newblock {\em Physica D: Nonlinear Phenomena}, vol. 31, no. 2, pp. 277--283,
  1988.

\bibitem{Ooi14AER}
C.~S. Ooi, K.~P. Seng, L.~M. Ang, and L.~W. Chew,
\newblock ``A new approach of audio emotion recognition,''
\newblock {\em Expert Syst. Appl.}, vol. 41, pp. 5858--5869, 2014.

\bibitem{Rubinstein98sparse}
R.~Rubinstein, A.~M. Bruckstein, and M.~Elad,
\newblock ``Dictionaries for sparse representation modeling,''
\newblock {\em Proc. IEEE}, vol. 98, no. 6, pp. 1045--1057, 2010.

\bibitem{Miao11MMC}
M.~Cheng, Y.~Y. Tang, and C.~M. Pun,
\newblock ``Nonparametric feature extraction via direct maximum margin
  alignment,''
\newblock in {\em Proceedings of Int. Conf. Mach. Learn. Appl.}, 2011.

\end{thebibliography}

\end{document}